\documentclass[letterpaper, 10 pt, journal, twoside]{IEEEtran}

\IEEEoverridecommandlockouts

\usepackage[english]{babel}
\usepackage{blindtext}
\usepackage{graphicx}
\usepackage{comment}
\usepackage{lipsum}
\usepackage{wrapfig}
\usepackage{booktabs}
\usepackage{subcaption}
\usepackage{titlesec}
\usepackage{caption}
\usepackage{textcomp}
\usepackage{hyperref}
\usepackage{cuted}
\usepackage{algorithm}
\usepackage{algpseudocode}
\usepackage{xcolor}
\usepackage[normalem]{ulem}

\begin{document}

\title{
Language Models as\\Zero-Shot Trajectory Generators
}

\author{Teyun Kwon$^{1}$, Norman Di Palo$^{1}$, Edward Johns$^{1}$
\thanks{Manuscript received: December 27, 2023; Revised April 12, 2024; Accepted May 13, 2024.}
\thanks{This paper was recommended for publication by Editor Aleksandra Faust upon evaluation of the Associate Editor and Reviewers' comments. This work was supported by the Royal Academy of Engineering under the Research Fellowship Scheme.} 
\thanks{$^{1}$
Teyun Kwon, Norman Di Palo and Edward Johns are
with the Robot Learning Lab at Imperial College London.
        {\tt\footnotesize \{john.kwon20, n.di-palo20, e.johns\}@imperial.ac.uk}}%
\thanks{Digital Object Identifier (DOI): 10.1109/LRA.2024.3410155.}
}

\markboth{IEEE Robotics and Automation Letters. Preprint Version. Accepted May, 2024}
{Kwon, Di Palo, Johns: Language Models as Zero-Shot Trajectory Generators
}

\maketitle

\begin{strip}
    \vspace{-75pt}
    \centering
    \includegraphics[width=\linewidth]{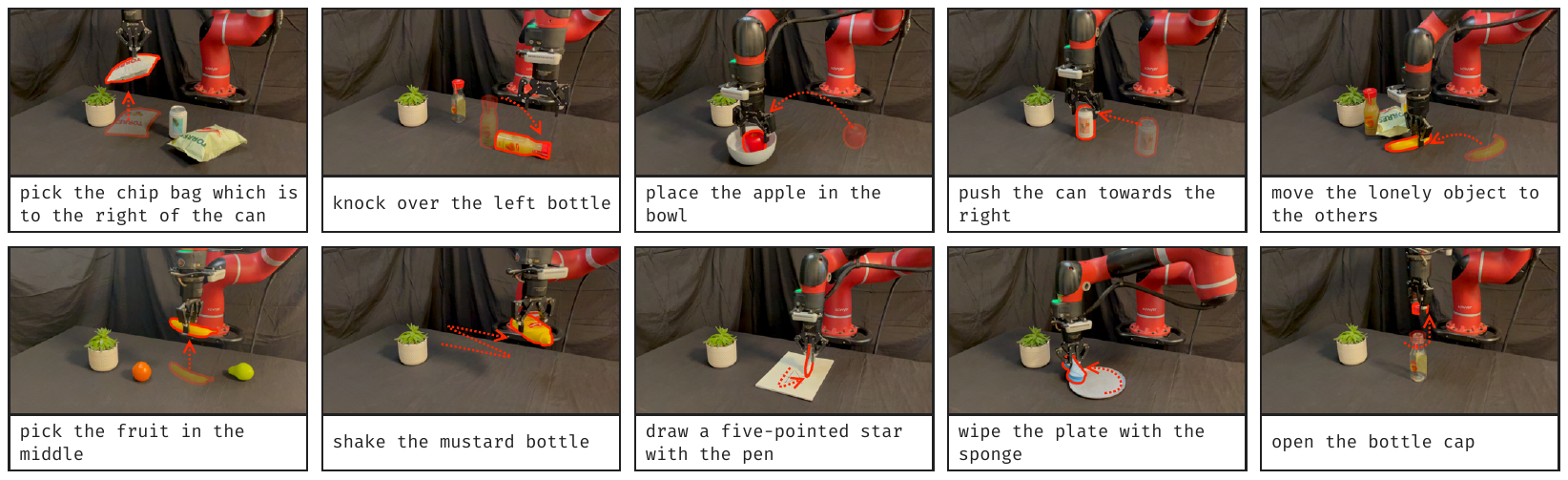}
    \captionof{figure}{A selection of the tasks we use to study if a single, task-agnostic LLM prompt can generate a dense sequence of end-effector poses, when given only object detection and segmentation models, and no in-context examples, motion primitives, pre-trained skills, or external trajectory optimisers.}
    \label{fig:tasks}
    \vspace{-5pt}
\end{strip}


\begin{abstract}
Large Language Models (LLMs) have recently shown promise as high-level planners for robots when given access to a selection of low-level skills. However, it is often assumed that LLMs do not possess sufficient knowledge to be used for the low-level trajectories themselves. In this work, we address this assumption thoroughly, and investigate if an LLM (GPT-4) can directly predict a dense sequence of end-effector poses for manipulation tasks, when given access to only object detection and segmentation vision models. We designed a single, task-agnostic prompt, without any in-context examples, motion primitives, or external trajectory optimisers. Then we studied how well it can perform across 30 real-world language-based tasks, such as ``\textit{open the bottle cap}" and ``\textit{wipe the plate with the sponge}", and we investigated which design choices in this prompt are the most important. Our conclusions raise the assumed limit of LLMs for robotics, and we reveal for the first time that LLMs do indeed possess an understanding of low-level robot control sufficient for a range of common tasks, and that they can additionally detect failures and then re-plan trajectories accordingly. Videos, prompts, and code are available at: \href{https://www.robot-learning.uk/language-models-trajectory-generators}{https://www.robot-learning.uk/language-models-trajectory-generators}.
\end{abstract}



\IEEEpeerreviewmaketitle

\section{Introduction}

\IEEEPARstart{I}{n} recent years, Large Language Models (LLMs) have attracted significant attention and acclaim for their remarkable capabilities in reasoning about common, everyday tasks~\cite{zhao2023survey}. This widespread recognition has since led to efforts in the robotics community to adopt LLMs for high-level task planning~\cite{wang2023survey}. However, for low-level control, existing proposals have relied on auxiliary components beyond the LLM, such as pre-trained skills, motion primitives, trajectory optimisers, and numerous language-based in-context examples (Fig.~\ref{fig:taxonomy}). Given the lack of exposure of LLMs to physical interaction data, it is often assumed that LLMs are incapable of low-level control~\cite{huang2023voxposer, yu2023language, ahn2022saycan}.

However, until now, this assumption has not been thoroughly examined. In this paper, we now investigate if LLMs have sufficient understanding of low-level control to be adopted for \textbf{zero-shot dense trajectory generation for robot manipulators}, without the need for the aforementioned auxiliary components. We provide an LLM (GPT-4~\cite{openai2023gpt4}) with access to off-the-shelf object detection and segmentation models, and then require all remaining reasoning to be performed by the LLM itself. We also require that the same task-agnostic prompt is used for all tasks, such as ``\textit{open the bottle cap}" and ``\textit{wipe the plate with the sponge}", which we took from the recent literature. And through this investigation, we uncovered the underlying principles and strategies that empower LLMs to navigate the complexities of robot manipulation.

Consequently, our contributions are threefold:
\textbf{(1)} We show, for the first time, that a pre-trained LLM, when provided with only an off-the-shelf object detection and segmentation model, can \textbf{guide zero-shot a robot manipulator by outputting a dense sequence of end-effector poses}, without the need for pre-trained skills, motion primitives, trajectory optimisers, or in-context examples.
\textbf{(2)} We present several ablation studies which shed light on \textbf{what techniques and prompts lead to the emergence of these capabilities}.
\textbf{(3)} We study how, by analysing the trajectory of objects across an image, \textbf{LLMs can also detect if a task has failed and subsequently re-plan an alternative trajectory}.

\section{Related Work} \label{sec:related}

Preliminary versions of this paper were accepted at the ICRA 2024 VLMNM Workshop (Spotlight), the CoRL 2023 Workshop on Language and Robot Learning, and the NeurIPS 2023 Robot Learning Workshop (all non-archival). This work now presents our full paper, with further experiments and analysis.

While prior works have made significant strides in leveraging LLMs for various aspects of robotic control~\cite{wang2023survey}, several limitations and dependencies on external modules persist. The core motivation of our work is to \textbf{investigate whether these limitations are inherent, or if LLMs can be deployed for low-level control}, going from language to a dense sequence of end-effector poses. In this section, we provide an overview of the relevant literature and highlight key distinctions between prior approaches and our research focus.

\begin{figure}
    \centering
    \includegraphics[width=\linewidth]{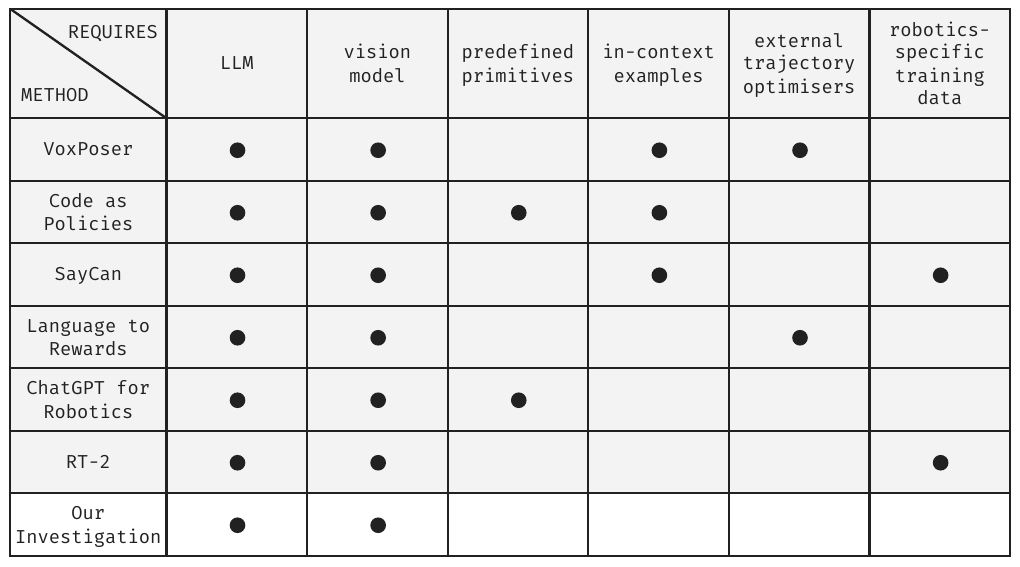}
    \caption{A taxonomy of requirements of LLM-based zero-shot methods from the recent literature.}
    \label{fig:taxonomy}
    \vspace{-12.5pt}
\end{figure}

\textbf{LLMs for Robotics:} There have been a number of works which leverage the common-sense knowledge and instruction-following capabilities of LLMs, but they have relied on external components for the full low-level trajectory generation. Both Code as Policies~\cite{liang2023code} and ChatGPT for Robotics~\cite{vemprala2023chatgpt} rely on predefined motion primitives for robot control, and their focus is predominantly on high-level planning. In the case of SayCan~\cite{ahn2022saycan}, robotics-specific data is required to pre-train the skills. VoxPoser~\cite{huang2023voxposer} and Language to Rewards~\cite{yu2023language} have explored the use of LLMs to generate high-reward regions for robot movement, but these methods still necessitate external trajectory optimisers to compute a trajectory, such as cost and reward functions used to evaluate randomly sampled trajectories along with Model Predictive Control (MPC)~\cite{huang2023voxposer}. VoxPoser~\cite{huang2023voxposer}, Code as Policies~\cite{liang2023code}, and SayCan~\cite{ahn2022saycan} have also relied heavily on providing in-context examples to the LLM input. However, these methods can encounter challenges when extrapolating beyond the demonstrated tasks. A summary of these works and their required auxiliary components is shown in Fig.~\ref{fig:taxonomy}.

Out of the aforementioned works, Code as Policies~\cite{liang2023code} is closest to our work. However, instead of relying on predefined primitives and in-context examples, we focus our investigation on the generation of trajectories more complex than a sequence of linear interpolations and without any task-specific guidance, thus broadening the scope of applicability and adaptability in the real world and reducing the reliance on human expertise. We also show that these complex trajectories are generated from the LLMs' internal knowledge of these tasks, instead of relying solely on the use of external libraries for code generation, and that this understanding is not only beneficial for trajectory generation, but also for task success detection as well, providing insight into the capabilities of LLMs to detect and recover from failures.

\textbf{Foundation Models for Robotics:} Recent works~\cite{brohan2023rt2,driess2023palme} demonstrated that a Vision Language Model (VLM)~\cite{alayrac2022flamingo} can be fine-tuned with a large robotics-related dataset of actions to enable zero-shot language-conditioned control. It has also been shown that such demonstrations can be provided directly to LLMs in the form of language tokens as in-context examples~\cite{KAT,general-pattern-machines}. Other works on foundation models include Socratic Models~\cite{socratic-models}, which proposes to leverage multiple pre-trained models via language-based exchange, and Inner Monologue~\cite{inner-monologue} which focuses on environment feedback for success detection, scene description, and human interaction. VLMs and LLMs have also been used in reinforcement learning settings for their planning and success detection capabilities, especially in long-horizon tasks~\cite{unified-agent}.

\textbf{Language and Visual Grounding:} Although not the main focus of our work, there have been numerous methods to ground the visual information for language-conditioned robot manipulation, such as embedding CLIP features into NeRFs~\cite{distilled-feature-fields,lerf,lerf-togo}, and keypoint extraction for object encoding~\cite{KAT} and demonstration retrieval~\cite{dinobot,retrieval-alignment-replay}. DALL-E-Bot~\cite{dallebot} introduces web-scale image diffusion models to ground the target rearrangement scene, whereas CLIP similarity scores are used in Dream2Real~\cite{dream2real} to determine the most visually plausible target scene.

\section{Problem Formulation}

We investigate if an LLM (GPT-4~\cite{openai2023gpt4}) can predict a dense sequence of end-effector poses to solve a range of manipulation tasks. We now explain what the assumptions and constraints are in our investigation, followed by details of our real-world experimental setup, and the tasks used for evaluation. Given this background, we then present our investigation and its results in Sec.~\ref{sec:prompt_development}.

\textbf{Assumptions and Constraints:} We design a task-agnostic prompt to study the zero-shot control capabilities of LLMs, with the following assumptions: \textbf{(1)} no pre-existing motion primitives, policies or trajectory optimisers: the LLM should output the \textit{full sequence of end-effector poses itself}; \textbf{(2)} no in-context examples: we study the ability of LLMs to reason about tasks given their \textit{internal knowledge alone}, and no part of any task is explicitly mentioned in the prompt, either in the form of examples or instructions; \textbf{(3)} the LLM can query a pre-trained vision model to obtain information about the scene, but should \textit{autonomously generate, parse and interpret the inputs and outputs}; \textbf{(4)} no additional pre-training or fine-tuning on robotics-specific data: we focus our research on \textit{pre-trained and widely available models}, so that our work can easily be reproduced even with limited compute budget.

\begin{figure}
    \centering
    \includegraphics[width=\linewidth]{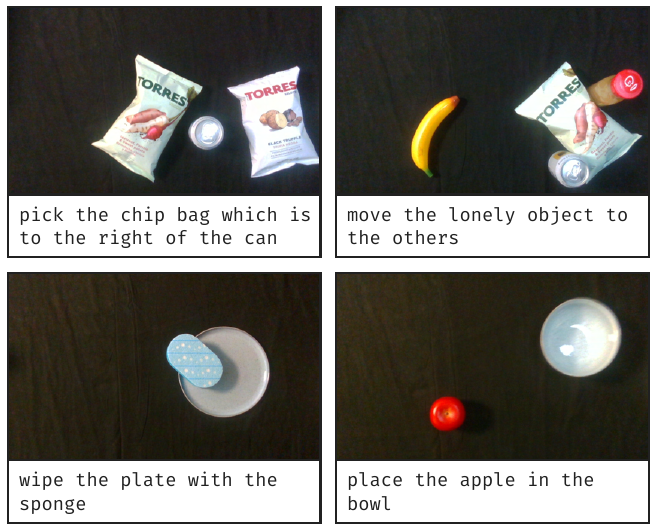}
    \caption{Example wrist-camera observations received by the robot at the start of each task, and their corresponding task instructions.}
    \label{fig:observations}
    \vspace{-15pt}
\end{figure}

\textbf{Real-World Experimental Setup:} We run our experiments on a Sawyer robot equipped with a 2F-85 Robotiq gripper. We use two Intel RealSense D435 RGB-D cameras, one mounted on the wrist of the robot, and the other fixed on a tripod, to observe the environment. The wrist-mounted camera captures a top-down view of the environment at the beginning of an episode (Fig.~\ref{fig:observations}), which is used by a vision model if queried by the LLM. We utilise a pre-trained object detection model, LangSAM~\cite{langsam} (based on Grounding DINO~\cite{liu2023grounding} and Segment Anything~\cite{kirillov2023segment}), and whenever the LLM calls \texttt{detect\_object}, we automatically calculate 3-D bounding boxes of the queried objects from the segmentation maps returned by LangSAM using the camera calibration, and provide the bounding boxes to the LLM. The LLM then leverages this visual understanding of the environment to predict a sequence of 4-D end-effector poses (3 dimensions for position, 1 dimension for rotation about the vertical axis), as well as either \texttt{open\_gripper} or \texttt{close\_gripper} commands. This is then executed by the robot in an open loop, using a position controller to move sequentially between each pose, hence producing a full trajectory. After task execution, we use XMem~\cite{cheng2022xmem} to track the segmentation maps over the entire duration of the task, which is then later used for detecting if the task was successful or not.

\textbf{Task Selection:} In pursuit of objectivity, we opt to benchmark our zero-shot LLM-guided robotic control against a challenging repertoire of everyday manipulation tasks. We \textbf{recreated 30 everyday manipulation tasks from recent robotics papers} published at leading conferences~\cite{ahn2022saycan, xiao2023dial, brohan2023rt1, yu2023rosie, huang2023voxposer}, often tackled by relying on hundreds of manual demonstrations. This serves as a representative benchmark of real-world challenges, mirroring the complexity and diversity of the tasks encountered in contemporary robotics research. We choose tasks which semantically cover the most representative tabletop robot behaviours expressed in these papers, and success criteria are human-evaluated and designed to mirror those proposed in the original papers. For each combination of task and method in the following experimental sections, we calculate the success rate over 5 trials, randomising the positions and orientations of the objects for each trial. The task description is provided in natural language to the LLM, after which no additional human feedback or intervention is allowed. The full list of tasks is shown in Fig.~\ref{fig:tasks_success}, and example task success and failure videos are available at: \href{https://www.robot-learning.uk/language-models-trajectory-generators}{https://www.robot-learning.uk/language-models-trajectory-generators}.

\begin{figure*}
    \centering
    \includegraphics[width=0.7\textwidth]{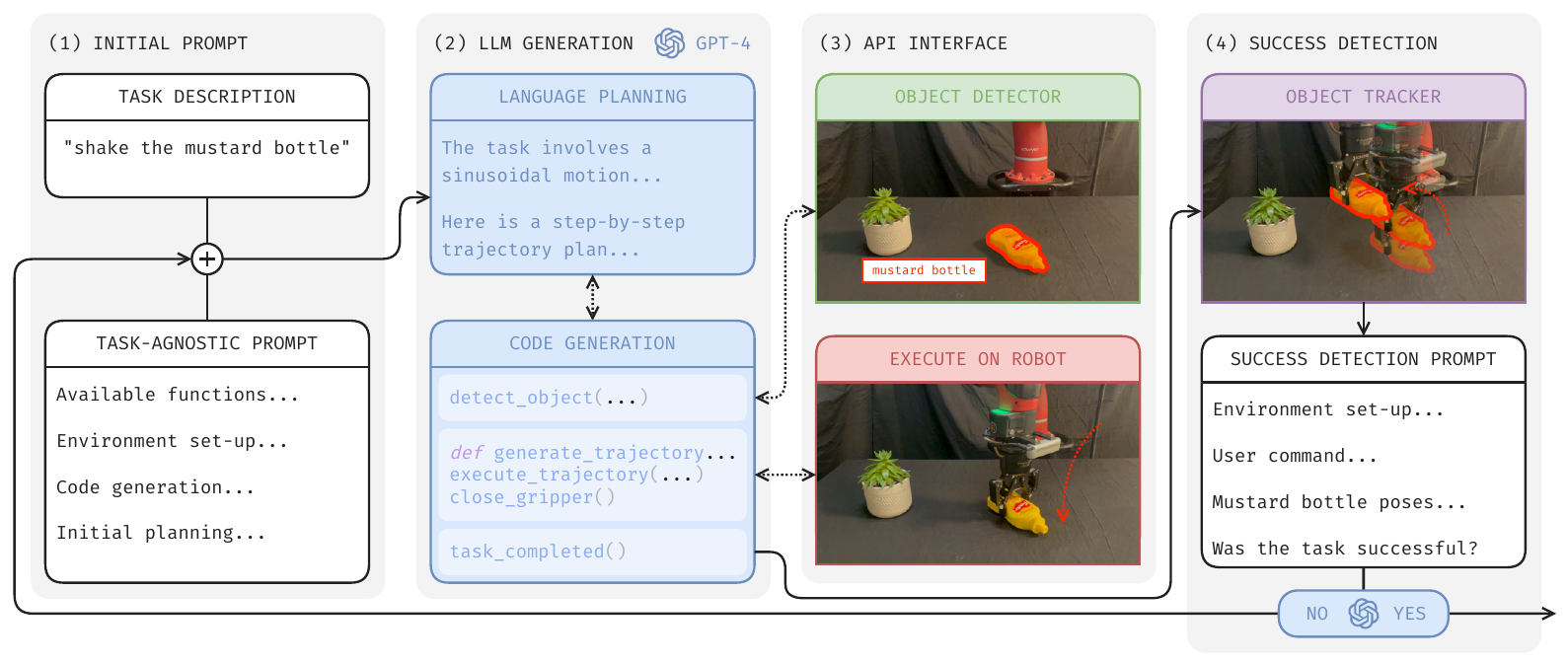}
    \caption{An overview of the pipeline. \textbf{(1)} The main prompt along with the task instruction is provided to the LLM, from which it \textbf{(2)} generates high-level natural language reasoning steps before outputting Python code \textbf{(3)} to interface with a pre-trained object detection model and execute the generated trajectories on the robot. \textbf{(4)} After task execution, an off-the-shelf object tracking model is used to obtain 3-D bounding boxes of the previously detected objects over the duration of the task, which are then provided to the LLM as numerical values to detect whether the task was executed successfully or not.}
    \label{fig:overview}
\end{figure*}

\section{Prompt Development} \label{sec:prompt_development}

\textbf{Full Prompt:} The core motivation of our work is to investigate whether LLMs can inherently guide robots with minimal dependence on specialised external models and components, in order to provide effective and useful insights for the robotics community. Through this investigation, we designed a single task-agnostic prompt for a range of everyday manipulation tasks, which does not require any in-context examples or task-specific guidance. Fig.~\ref{fig:overview} illustrates the main information flow in our framework, showing how the task-agnostic prompt interfaces with the vision models and the robot.

\begin{figure*}[t]
    \centering
    \includegraphics[width=0.7\textwidth]{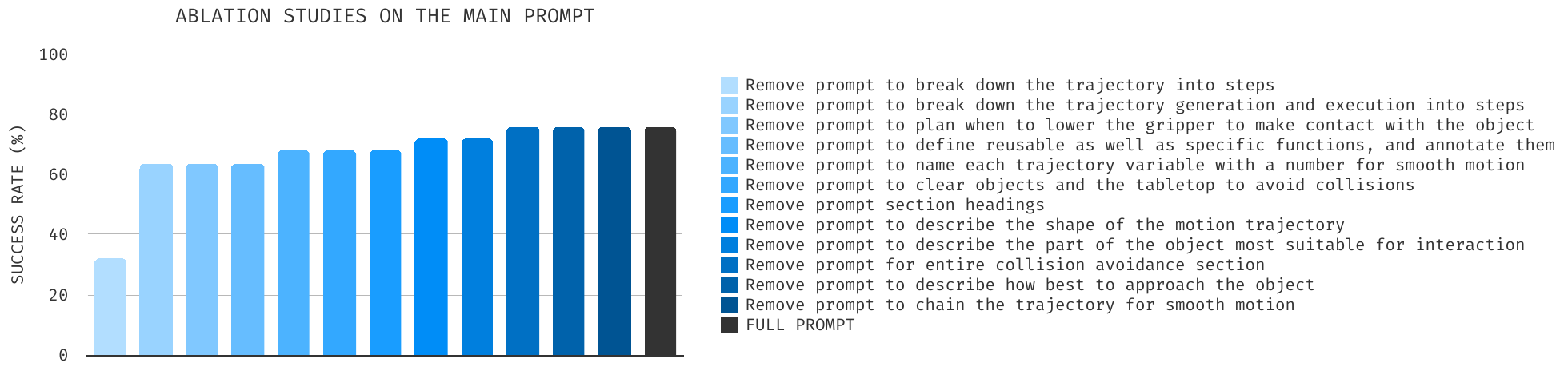}
    \caption{We investigate the effect of removing parts of the main prompt on task success rates.}
    \label{fig:remove-prompt}
    \vspace{-10pt}
\end{figure*}

Through our experiments outlined in this section, our final prompt formulation instructs the LLM to self-summarise and decompose the predicted plan into steps, before generating Python code which, when run by a standard Python interpreter, outputs a dense sequence of poses for the end-effector to follow; this pipeline resulted in the best performance across those we experimented with. We include details fundamental to all tasks, such as coordinate definitions, as well as functions available for the LLM to call, such as \texttt{detect\_object}, which returns the calculated 3-D bounding boxes of the queried objects directly to the LLM. We also include instructions which aim to improve the correctness and reliability of the generated trajectories, such as guidance on step-by-step reasoning, code generation, and collision avoidance.

\textbf{Prompt Ablations:} During the design of this full prompt, we identified several challenges when using LLMs for low-level control, without access to other external dependencies. In this section, we now outline these challenges which motivated the final design of the prompt, and accompany them with results from ablation studies conducted across a diverse set of tasks (Fig.~\ref{fig:remove-prompt}), where certain parts of the full prompt were removed.

\textbf{(1) LLMs often require step-by-step reasoning to solve tasks.} Prior work has shown that the reasoning capabilities of LLMs can be improved by asking them to break down the task in a step-by-step manner~\cite{wei2023chainofthought, kojima2023stepbystep}. Adopting this strategy, we prompt the LLM \textbf{(1)} to break down the trajectory into a sequence of sub-trajectory steps, and \textbf{(2)} to include in the plan when to lower the gripper to make contact with an object. We find that, without including these step-by-step reasoning prompts, the LLM often omits key trajectory steps required to execute the task successfully, such as opening or closing the gripper, or aligning the gripper to be parallel to the graspable side of the object, which are not stated explicitly in the prompt. Indeed, the first three columns in Fig.~\ref{fig:remove-prompt} show that prompting the LLM to think step by step resulted in the highest performance increase.

\textbf{(2) LLMs can be prone to write code which results in errors, both syntactically and semantically.} While much improvement has been made in the domain of code generation by LLMs~\cite{chen2021evaluating, openai2023gpt4}, their outputs can still throw errors, as well as produce undesirable results when executed. In order to mitigate this, and inspired again by the power of LLMs performing an internal monologue with natural language reasoning, we prompt the LLM to document any functions it defines, with their expected inputs and outputs, and their data types. In addition, we include a prompt instructing the LLM to define reusable functions for common motions (for example, linear trajectory from one point to another), to prevent instances where, as a notable example, it would hard-code the height of the gripper inside a function definition, and reuse that function for another sub-trajectory step which should have been executed at a different height. Similarly, we prompt the LLM to name each sub-trajectory step variable with a number to relate it to each of the steps in the high-level trajectory plan, and to minimise the chance of omitting a sub-trajectory step. The effects of removing these prompt components are, again, noticeable (fourth and fifth columns in Fig.~\ref{fig:remove-prompt}).

\begin{figure*}
    \centering
    \includegraphics[width=0.7\textwidth]{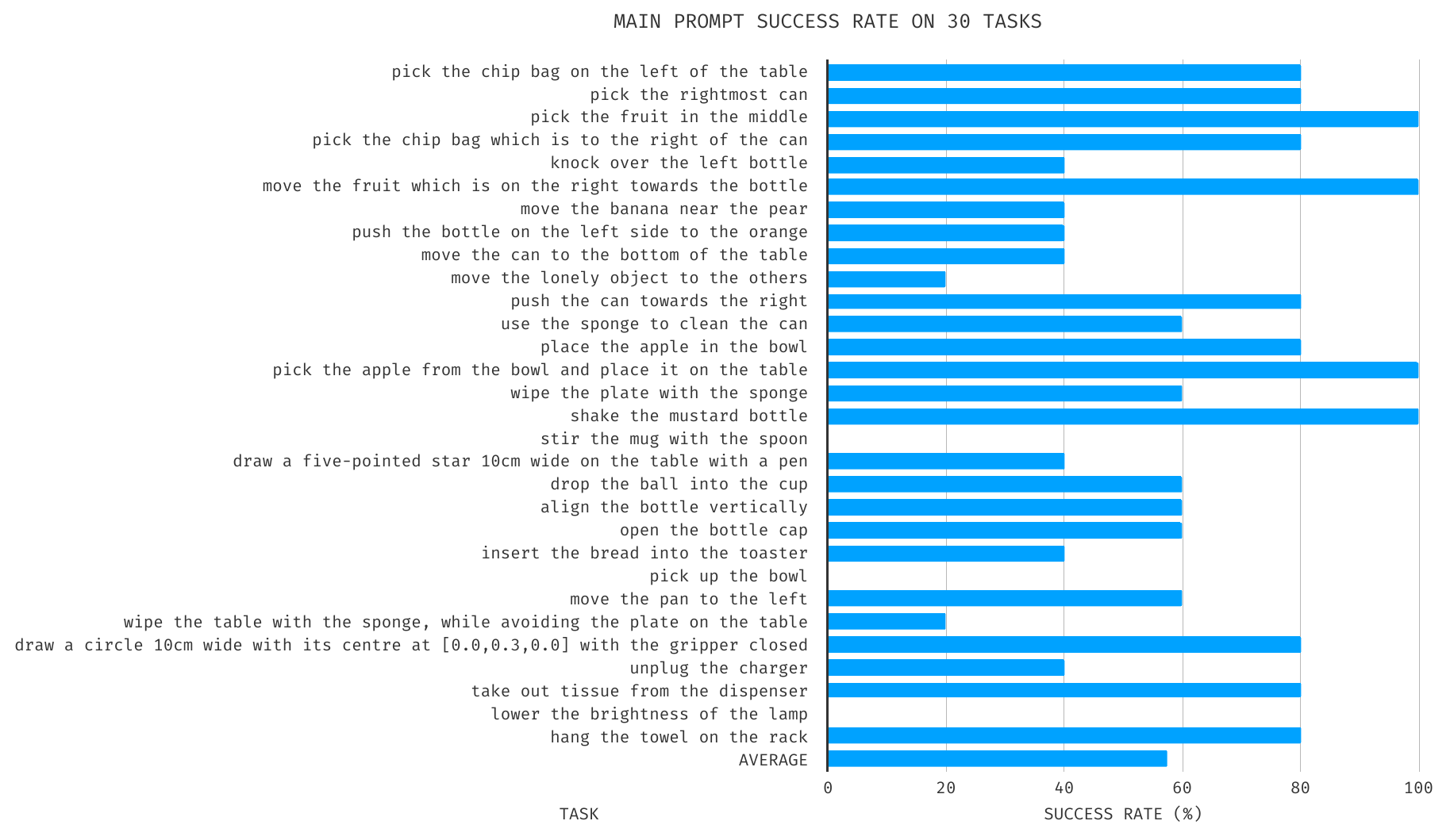}
    \caption{Success rates of the full prompt for 30 manipulation tasks.}
    \label{fig:tasks_success}
    \vspace{-10pt}
\end{figure*}

\textbf{(3) LLMs are trained on limited grounded physical interaction data.} Due to the scarcity of grounded physical interaction data in their training corpora~\cite{chinchilla}, LLMs often fail to take into account possible collisions between the objects being manipulated. We therefore prompt the LLM to pay attention to the dimensions of the objects and to generate additional waypoints and sub-trajectories, which could help with avoiding collisions. We also include in the prompt a specific phrase which we noticed during our investigation was being used frequently by the LLM for its internal reasoning (``\textit{clear objects and the tabletop}"). Our experiments show that, while removing this particular phrase from the collision avoidance prompt lowered performance (sixth column in Fig.~\ref{fig:remove-prompt}), LLMs do possess some inherent understanding of possible collisions between different objects, as they performed well even after removing the entire collision avoidance prompt (tenth column in Fig.~\ref{fig:remove-prompt}).

\textbf{(4) LLMs often fail to reason about complex trajectory shapes.} In a manner similar to Challenge (1), we employ a two-step strategy, where initially, we explicitly ask the LLM to generate a textual description of the \textit{shape of the motion trajectory} as internal reasoning (for example, shaking involves a sinusoidal motion), before outputting the actual sequence of poses required to execute this trajectory (however, the difference is that in Challenge (1), we prompted the LLM to output a more detailed step-by-step trajectory plan). This has been shown to be beneficial in prior work~\cite{yu2023language}, and indeed this result is also reflected in the eighth column in Fig.~\ref{fig:remove-prompt}.

\textbf{(5) LLMs often fail to reason about how to interact with objects.} In our experiments, we found that LLMs often simplified and failed to reason about more intricate details of object interaction, such as realising that some objects require interaction with a specific part (for example, the rim of a bowl, or the handle of a pan). In order to enable the LLM to detect the most suitable object part to interact with, we prompt it to describe the object part in high-level natural language, and we see in the ninth column in Fig.~\ref{fig:remove-prompt} that this results in more tasks being executed successfully.

\textbf{Full Prompt Evaluation:} Here, we now investigate the LLM's ability to solve zero-shot a range of manipulation tasks, by evaluating the full prompt on the full set of tasks taken from the recent literature. These tasks and their success rates are presented in Fig.~\ref{fig:tasks_success}. Remarkably, our experiments reveal that LLMs, when equipped with an off-the-shelf vision model and no external motion primitives, policies, or trajectory optimisers, do indeed exhibit notable proficiency in executing the majority of these tasks, by directly predicting a dense sequence of end-effector poses. In the original papers from which these tasks are selected~\cite{brohan2023rt1, xiao2023dial, yu2023rosie}, solving these tasks required numerous human demonstrations. As such, these findings underscore the potential of LLMs as intuitive and versatile guides for robotic manipulation that minimise the need for human time and supervision. The full LLM prompts, together with sample LLM outputs, are available at: \href{https://www.robot-learning.uk/language-models-trajectory-generators}{https://www.robot-learning.uk/language-models-trajectory-generators}.

\section{Further Investigations} \label{sec:further-investigations}

\begin{wrapfigure}{L}{0.5\linewidth}
    \vspace{-12pt}
    \centering
    \includegraphics[width=\linewidth]{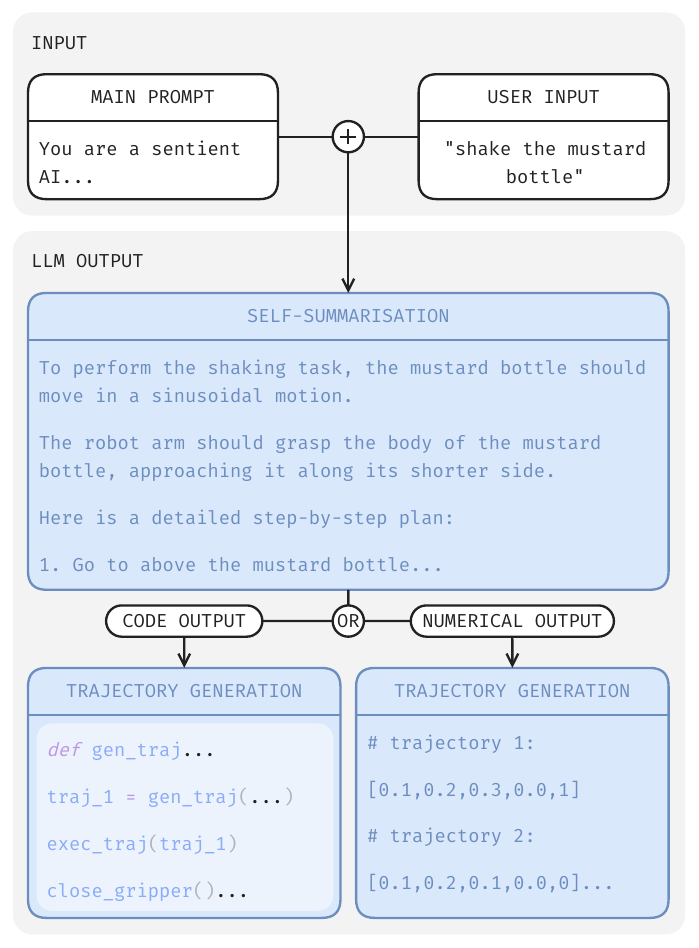}
    \caption{Given the full main prompt and the user input command, the LLM first outputs a high-level natural language self-summarisation of the trajectory plan, before generating either code which computes and executes the trajectory, or the final trajectory directly as a list of numerical values.}
    \label{fig:llm-reasoning-steps}
\end{wrapfigure}

In this section, we detail further ablation studies conducted regarding the modality of the trajectory generation, the extent to which each output modality is executable by the robot, the performance of the different available LLMs, and the ability of LLMs to detect whether a task was executed successfully or not and subsequently re-plan the trajectory. We also detail the various sources of error which led to task execution failures when performing the 30 tasks with the main prompt, and provide a detailed comparison against Code as Policies~\cite{liang2023code}, which is closest to our work.

\begin{figure*}
    \centering
    \includegraphics[width=\textwidth]{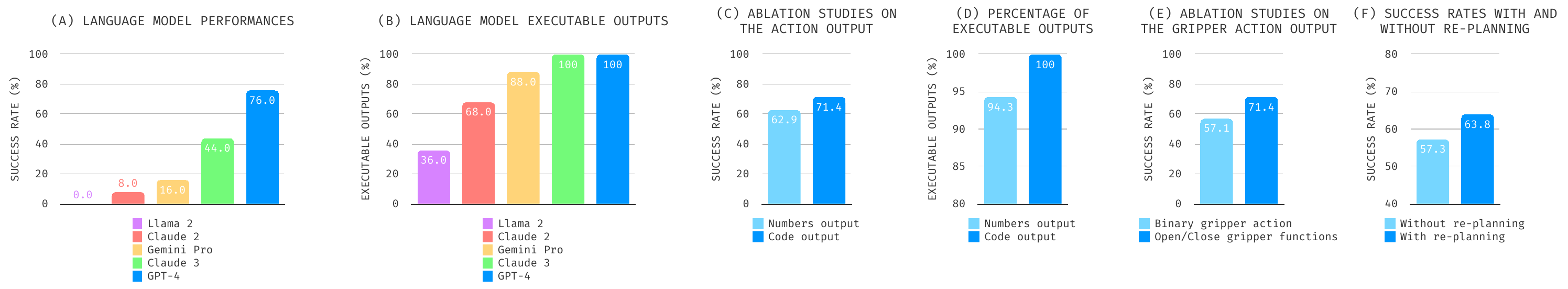}
    \caption{\textbf{(A)} We compare the performance of different widely used LLMs. \textbf{(B)} We compare the abilities of these LLMs to generate outputs that are directly executable by the robot. \textbf{(C)} We compare different modes for the trajectory output. \textbf{(D)} We measure the percentage of control outputs from the LLM that are directly executable by the robot. \textbf{(E)} We compare different modes for controlling the gripper. \textbf{(F)} We demonstrate the ability of LLMs to detect failures and re-plan autonomously.}
    \label{fig:many_exps}
    \label{fig:numbers_code}
    \label{fig:replanning_results}
    \label{fig:gripper_action}
    \label{fig:llms}
    \vspace{-14pt}
\end{figure*}

\textbf{(1) How should the final trajectory be represented?} In this set of experiments, we explore the optimal way for the LLM to output the sequence of end-effector poses. Specifically, we conduct ablation studies to evaluate whether this should be represented as a list of \textit{numerical values} or as \textit{code for trajectory generation}. Fig.~\ref{fig:llm-reasoning-steps} shows the distinction between these two output modes.

The results, summarised in Fig.~\ref{fig:many_exps} C, offer valuable insights. Notably, our investigation shows that outputting code that generates the trajectory outperforms predicting the trajectory directly as an explicit list of numerical poses for the end-effector to follow, represented as language tokens (Fig.~\ref{fig:llm-reasoning-steps}). In particular, we observe that representing trajectories as numerical values or as code yields similar performance for most tasks, with distinctions emerging in cases involving more intricate trajectories (for example, drawing a circle or a five-pointed star), where outputting code that generates such trajectories prevails (60\% success rates for code output compared to 10\% for numerical output). This suggests a \textit{fundamental property of LLMs for control}: while not trained on physical interactions and trajectories, their \textbf{understanding of both code, and mathematical and geometrical structures}~\cite{openai2023gpt4, luo2023wizardmath} \textbf{can bridge these two modes of thinking}. Once the overall trajectory shape has been identified by the LLM, while it can be challenging to follow it directly in numbers, it is proficient at generating code that itself can follow complex paths.

Additionally, we study whether directly generating a list of numerical poses, or code that then generates the poses itself, leads to executable outputs more often. Giving low-level control to the LLM poses the risk of the robot receiving wrongly formatted outputs that cannot be executed by the robot. Therefore, in this ablation, we investigate how often the output of the LLM is formatted such that it is executable by the robot. We include prompts instructing the LLM to follow a specific format for the trajectory generation (for the numerical poses, we require a list between the \texttt{\textlangle trajectory\textrangle} and \texttt{\textlangle /trajectory\textrangle} tags \textit{without any Python functions}, and for the code, we require any Python code to be between the \texttt{\textasciigrave\textasciigrave\textasciigrave python} and \texttt{\textasciigrave\textasciigrave\textasciigrave} tags). Given the output of the LLM, if an error is thrown during automatic parsing according to this format, we provide the LLM with the error message and ask it to correct the output, for up to three times. Measuring the percentage of executable outputs (Fig.~\ref{fig:many_exps} D) demonstrates that outputting code results in 100\% of executable trajectories, while direct numerical values cannot be parsed even after three self-corrections for some episodes.

\textbf{(2) How should the LLM output the gripper action?} We also investigate the optimal way of letting the LLM \textit{control the gripper open or close action}: we compare using a binary variable $a \in \{0,1\}$ or explicit functions \texttt{open\_gripper} and \texttt{close\_gripper}. Our results, in Fig.~\ref{fig:gripper_action} E, demonstrate that the LLM achieves better performance when using explicit functions, while using a binary variable leads to more errors. A notable failure case stemmed from the LLM hard-coding the gripper state to be open in one of the functions it defined for itself, such that when the same function was then used to generate the object approach-and-lift sub-trajectory steps, the gripper failed to close and grasp the object. Having explicit functions to open and close the gripper, on the other hand, allowed a decoupling of these fundamental actions and enabled the correct functions to be called at any time during the overall trajectory generation plan.

\textbf{(3) Which of the currently available LLMs performs best at following instructions outlined in the prompt?} Next, we present results on ablation studies conducted to determine the best-performing LLM on the main prompt. We see in Fig.~\ref{fig:many_exps} B that out of the five most popular LLMs currently available for public use to date~\cite{openai2023gpt4,anthropic2023claude2,touvron2023llama,claude3,gemini}, and under the same constraints as the executable output study presented in Fig.~\ref{fig:many_exps} D, only GPT-4 and Claude 3 (Opus) were able to generate code which was able to be executed 100\% of the time (regardless of whether the resulting code output completed the task successfully or not), whereas the outputs for Gemini 1.0 Pro, Claude 2 and Llama 2 (70B Chat) were only executable 88.0\%, 68.0\% and 36.0\% of the time, respectively. The corresponding success rates for the five LLMs are shown in Fig.~\ref{fig:many_exps} A.

\begin{figure}
    \centering
    \includegraphics[width=\linewidth]{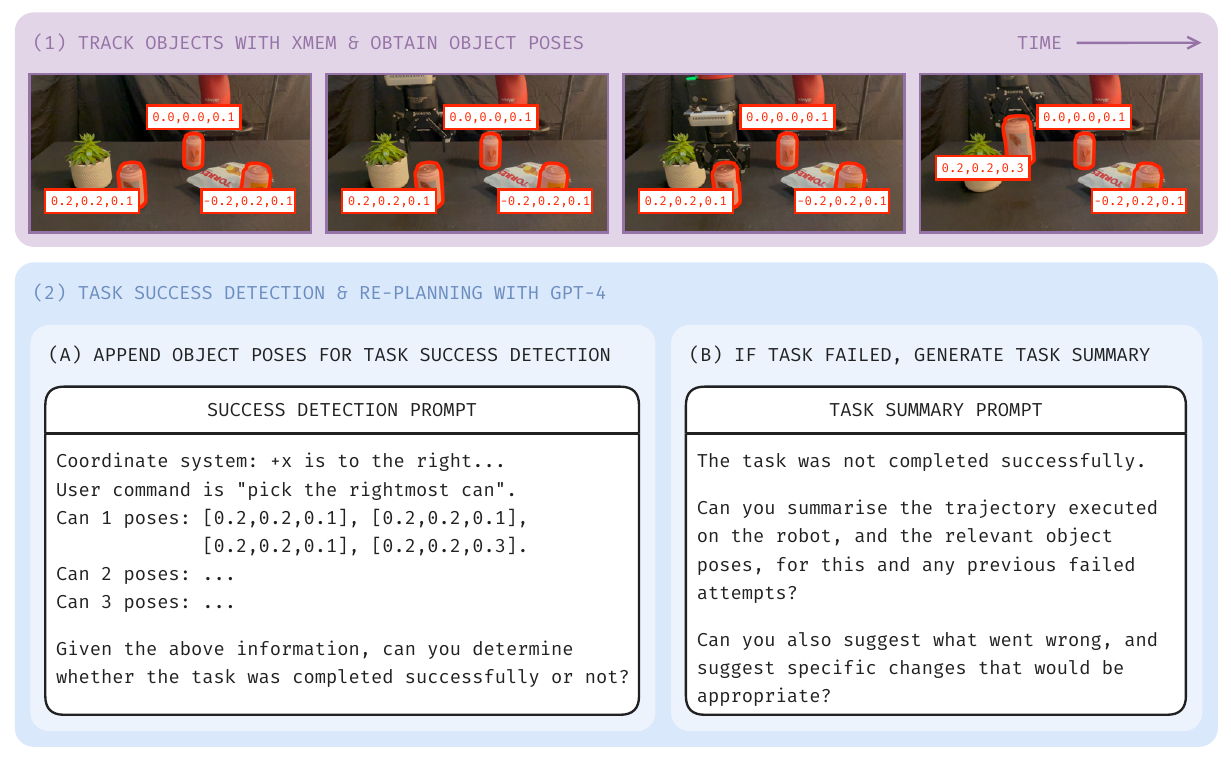}
    \caption{Our experiments demonstrate that LLMs can interpret the trajectories of objects to detect successful and unsuccessful episodes.}
    \label{fig:replan}
    \vspace{-10pt}
\end{figure}

\begin{wrapfigure}{L}{0.5\linewidth}
    \vspace{-12pt}
    \centering
    \includegraphics[width=\linewidth]{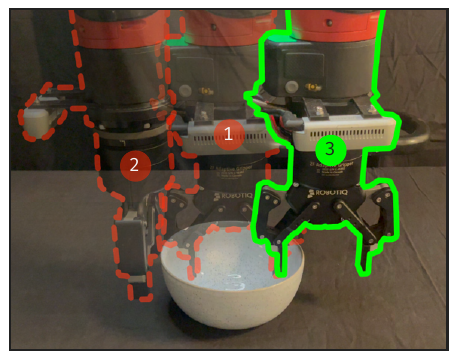}
    \caption{\textbf{(1)} The LLM attempts to grasp the bowl at its centroid, recognises failure, and \textbf{(2)} proposes a new trajectory. \textbf{(3)} On its third attempt after re-planning again, it successfully grasps the bowl.}
    \label{fig:bowl}
    \vspace{-15pt}
\end{wrapfigure}

\textbf{(4) Can LLMs recognise unsuccessful trajectories, and adapt their plan?} We also delve into the ability of LLMs to recognise and respond to failures during task execution. Our experiments demonstrate that, by analysing the numerical trajectories of objects, LLMs can autonomously detect failure outcomes and initiate re-planning to rectify them. We therefore demonstrate that LLMs possess not only the ability to generate trajectories, but also to \textbf{discern whether they represent successful or unsuccessful episodes, given the tasks requested by the user}. Our proposed pipeline for task success detection and re-planning is shown in Fig.~\ref{fig:replan}.

For each of the 5 trials of a task, when a failure is identified, the LLM modifies the original plan to tackle the possible issue. In Fig.~\ref{fig:many_exps} F, we demonstrate that this leads to a small improvement in performance, without the need for any human intervention. As a notable example, the LLM always fails at grasping a bowl on its first try (Fig.~\ref{fig:tasks_success}), attempting to grasp it by the centroid (Fig.~\ref{fig:bowl}). Through a sequence of two re-planning iterations, however, the LLM adapts its trajectory and then successfully grasps the bowl by its rim, leading to an increase from 0\% to 20\% in the overall task execution success rate.

\begin{figure}
    \centering
    \includegraphics[width=0.8\linewidth]{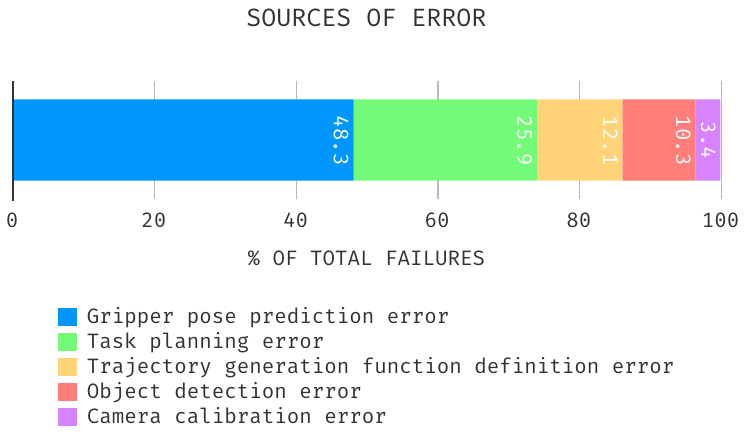}
    \caption{A breakdown of the different error types which caused task execution failures with the main prompt across the 30 tasks.}
    \label{fig:error-chart}
\end{figure}

\textbf{(5) What were the main failure modes?} We present the main failure modes of the main prompt on the 30 manipulation tasks. We group the sources of error into the following categories: \textbf{(1)} gripper pose prediction error, where the LLM was unable to predict accurate enough goal poses for the gripper, and this resulted in wrong parameterisations of trajectory generation functions (which the LLM defined for itself previously); \textbf{(2)} task planning error, where the LLM was unable to plan a correct high-level sequence of steps to complete the task successfully (for example, calling \texttt{close\_gripper} before lowering the gripper to the object to grasp it); \textbf{(3)} trajectory generation function definition error, where the LLM coded wrongly the function which would be used to generate the dense sequence of end-effector poses (the function itself would be executable by a Python interpreter, but as opposed to Error (1) which was concerned with the parameterisation of such functions, the function itself was wrongly defined; for example, hard-coding the gripper to be open or closed, or its height within the function definition); \textbf{(4)} object detection error, where the wrong segmentation map was returned by the object detector based on the text queried by the LLM; \textbf{(5)} camera calibration error, where the LLM might have generated the correct trajectory for the bounding box it was provided with, but the bounding box itself was incorrectly calculated due to noisy camera data and this resulted in the task being unsuccessful. The full breakdown is shown in Fig.~\ref{fig:error-chart}.

We can see that nearly half of the failure cases can be attributed to the more difficult task of predicting accurate gripper poses (48.3\%), compared to high-level planning (25.9\%), of which numerous works have already shown that LLMs are capable~\cite{ahn2022saycan,huang2022language}. Regarding the object detection errors, most of the errors were due to the object detector not being able to segment the object parts correctly when queried by the LLM, such as the rim of a bowl, or the slot of a toaster.

\begin{figure}
    \centering
    \includegraphics[width=0.8\linewidth]{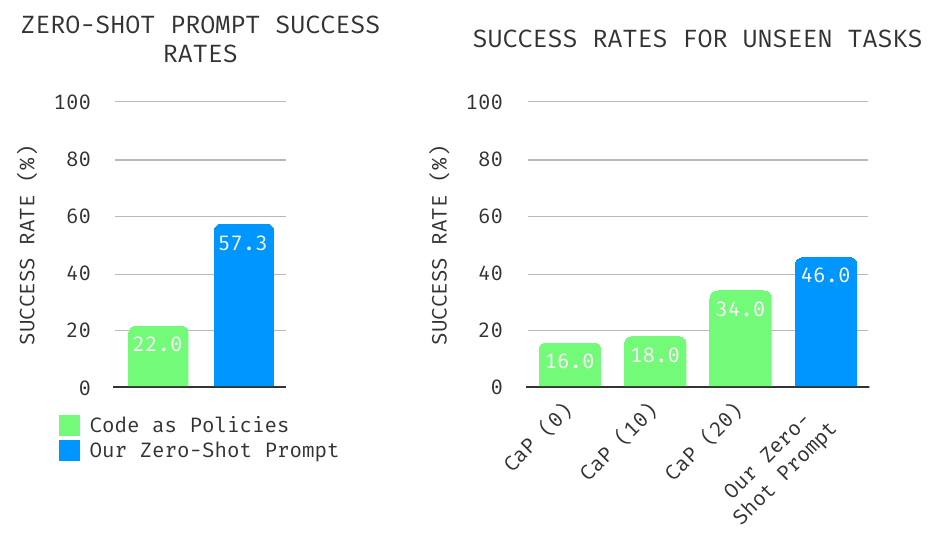}
    \caption{\textbf{(Left)} We compare our zero-shot task performance across 30 tasks to that of Code as Policies (CaP). \textbf{(Right)} We demonstrate that our prompt is able to generalise to unseen tasks with higher performance than using CaP-style prompting with a varying number of in-context examples (in brackets).}
    \label{fig:cap-results}
    \vspace{-3pt}
\end{figure}

\textbf{(6) How does our work compare to Code as Policies?} Finally, we perform two further sets of experiments in simulation and with ground-truth object poses, comparing our zero-shot task-agnostic prompt to the prompt structure set out in Code as Policies (CaP)~\cite{liang2023code}, which makes use of in-context examples and motion primitives. Firstly, we compare the performance of the original CaP prompt (containing 77 in-context examples) on the 30 tasks, using the original CaP API code, and compare to our zero-shot prompt. We find that while for basic pick-and-place tasks, for which examples had already been provided in the original CaP prompt, it has near-perfect success rates, it is unable to generalise at all to more complex tasks, resulting in a lower average success rate of 22.0\%, compared to 57.3\% for our prompt. These results are shown in Fig.~\ref{fig:cap-results} (Left).

Secondly, we study the role of CaP-style in-context examples in enabling (or limiting) unseen task generalisation, and compare on 10 unseen tasks the success rates of CaP prompts with a varying number of additional manually engineered in-context examples (0, 10, and 20). For the unseen tasks, we select 10 from the same set of 30 tasks, and for the additional in-context examples, we write one example for each of the remaining 20 tasks, and select 0, 10, or all 20 of them. We write these additional examples in the style of CaP, but covering a wider range of tasks beyond simple pick-and-place tasks, and append them to the original CaP prompt. We find that with these additional examples, the CaP prompt is able to generalise better to unseen tasks, but even with the 20 additional manually engineered in-context examples (which cover tasks of similar complexity to the 10 unseen tasks), the three CaP prompt variants are unable to perform as well (highest 34.0\%) as our zero-shot task-agnostic prompt (46.0\%). These results are shown in Fig.~\ref{fig:cap-results} (Right).

\section{Conclusions} \label{sec:conclusion}

The primary contribution of this paper is an investigation into to what extent an LLM can successfully predict dense sequences of end-effector poses for a range of real-world manipulation tasks. To differentiate this from other previous works, we imposed constraints that the LLM must use a single task-agnostic prompt without any in-context examples, and has access to only off-the-shelf object detection and segmentation vision models, with no other auxiliary components. Our experiments encompassed 30 diverse tasks drawn from the recent literature, and we showed that GPT-4, together with the prompt we designed, can perform well on many of these tasks.

With today's LLMs and our zero-shot prompt, there are several types of tasks which are too challenging, such as those requiring high precision, complex trajectory shapes, and richer perception beyond just computing bounding boxes. As LLMs continue to improve, we may see improved abilities to solve these tasks, and recent advances in VLMs will help to better interface language and vision. But for now, this paper raises the assumed limit of the utility of LLMs for robotics, and we hope that the insights we provide will help those developing their own LLM-based robot controllers, and will encourage those leading the field in LLMs and VLMs to further align developments with robotics applications.


\section*{Acknowledgement}

The authors wish to thank Kamil Dreczkowski, Georgios Papagiannis and Pietro Vitiello for their valuable discussion and feedback during the writing of the paper.


\bibliographystyle{IEEEtran}
\bibliography{IEEEabrv,RA-L/main}


\appendix

\section*{Pipeline Algorithm}

\begin{algorithm}
\textbf{Given:} full task-agnostic prompt $p$, and task instruction $l$
\caption{Zero-Shot Trajectory Generation with LLMs}\label{alg:method}
\begin{algorithmic}[1]
\State $prompt \gets p \oplus l$ \Comment{concatenate}
\State $task\_completed \gets$ False
\While{$task\_completed =$ False}
    \State $output \gets$ LLM($prompt$)
    \State $prompt \gets$ None
    \If{$output$ contains code} \Comment{extract code}
        \State \textbf{try} exec($output$)
        \State \textbf{except} Exception \textbf{then}
            \State \hspace{\algorithmicindent} $prompt \gets$ error message
        \State \textbf{else} \Comment{predefined APIs}
            \State \hspace{\algorithmicindent} \textbf{if} detect\_object($object$) is called \textbf{then}
                \State \hspace{\algorithmicindent}\hspace{\algorithmicindent} $pos, orn, dim \gets$ detect\_object($object$)
                \State \hspace{\algorithmicindent}\hspace{\algorithmicindent} $prompt \gets pos, orn, dim$
            \State \hspace{\algorithmicindent} \textbf{else if} execute\_trajectory($t$) is called \textbf{then}
                \State \hspace{\algorithmicindent}\hspace{\algorithmicindent} move\_robot($t$) \Comment{$t$ is list of poses}
            \State \hspace{\algorithmicindent} \textbf{else if} task\_completed() is called \textbf{then}
                \State \hspace{\algorithmicindent}\hspace{\algorithmicindent} $obj\_poses \gets$ XMem($traj\_frames$)
                \State \hspace{\algorithmicindent}\hspace{\algorithmicindent} $task\_completed \gets$ LLM($obj\_poses$)
                \State \hspace{\algorithmicindent}\hspace{\algorithmicindent} \textbf{if} $task\_completed =$ False \textbf{then}
                    \State \hspace{\algorithmicindent}\hspace{\algorithmicindent}\hspace{\algorithmicindent} reset\_environment()
                    \State \hspace{\algorithmicindent}\hspace{\algorithmicindent}\hspace{\algorithmicindent} $summary \gets$ LLM($obj\_poses$)
                    \State \hspace{\algorithmicindent}\hspace{\algorithmicindent}\hspace{\algorithmicindent} $prompt \gets p \oplus l \oplus summary$
                \State \hspace{\algorithmicindent}\hspace{\algorithmicindent} \textbf{end if}
            \State \hspace{\algorithmicindent} \textbf{end if}
        \State \textbf{end try}
    \EndIf
\EndWhile
\end{algorithmic}
\end{algorithm}

\section*{Tasks Selected for Ablation Studies}

\subsection{Ablation Studies on the Main Prompt \& Language Model Comparisons}

\begin{itemize}
    \item pick the chip bag which is to the right of the can
    \item place the apple in the bowl
    \item shake the mustard bottle
    \item open the bottle cap
    \item move the pan to the left
\end{itemize}

\subsection{Ablation Studies on the Action Output}

\begin{itemize}
    \item pick the chip bag which is to the right of the can
    \item place the apple in the bowl
    \item shake the mustard bottle
    \item open the bottle cap
    \item move the pan to the left
    \item draw a five-pointed star 10cm wide on the table with a pen
    \item draw a circle 10cm wide with its centre at [0.0,0.3,0.0] with the gripper closed
\end{itemize}

\subsection{Code as Policies Comparison}

\begin{itemize}
    \item open the bottle cap
    \item insert the bread into the toaster
    \item pick up the bowl
    \item move the pan to the left
    \item wipe the table with the sponge, while avoiding the plate on the table
    \item draw a circle 10cm wide with its centre at [0.0,0.3,0.0] with the gripper closed
    \item unplug the charger
    \item take out tissue from the dispenser
    \item lower the brightness of the lamp
    \item hang the towel on the rack
\end{itemize}

\addtolength{\textheight}{-17cm}

\vspace{5pt}
\section*{Task Success Criteria}

\begin{itemize}
    \item \textit{pick the chip bag on the left of the table}: the left chip bag is lifted at least 10cm off the table.
    \item \textit{pick the rightmost can}: the rightmost can is lifted at least 10cm off the table.
    \item \textit{pick the fruit in the middle}: the horizontally middle fruit is lifted at least 10cm off the table.
    \item \textit{pick the chip bag which is to the right of the can}: the chip bag to the right of the can is lifted at least 10cm off the table.
    \item \textit{knock over the left bottle}: the left bottle is knocked over so that it is lying on its side.
    \item \textit{move the fruit which is on the right towards the bottle}: the right fruit is placed within 5cm of the bottle.
    \item \textit{move the banana near the pear}: the banana is placed within 5cm of the pear.
    \item \textit{push the bottle on the left side to the orange}: the left bottle is pushed (remains in contact with the table throughout) within 5cm of the orange.
    \item \textit{move the can to the bottom of the table}: the can is placed within 10cm of the bottom edge of the table.
    \item \textit{move the lonely object to the others}: the object which is furthest away is placed within 5cm of one of the other objects.
    \item \textit{push the can towards the right}: the can is pushed (remains in contact with the table throughout) to the right by at least 10cm.
    \item \textit{use the sponge to clean the can}: the sponge touches the can.
    \item \textit{place the apple in the bowl}: the apple is placed inside the bowl.
    \item \textit{pick the apple from the bowl and place it on the table}: the apple is placed anywhere on the table out of the bowl.
    \item \textit{wipe the plate with the sponge}: the sponge is moved in any wiping motion (zigzag, circular, etc.) while remaining in contact with the plate throughout.
    \item \textit{shake the mustard bottle}: the mustard bottle is moved in any shaking motion (up-and-down, left-and-right, etc.) at least twice, at any speed.
    \item \textit{stir the mug with the spoon}: the spoon is picked up from an upright pose, inserted into the mug, and is moved in a circular motion inside the mug.
    \item \textit{draw a five-pointed star 10cm wide on the table with a pen}: the pen is picked up from an upright pose, and is moved while remaining in contact with the table throughout such that it draws a five-pointed star 10cm wide.
    \item \textit{drop the ball into the cup}: the ball is placed inside the cup.
    \item \textit{align the bottle vertically}: the bottle is rotated such that it is pointing either the top or bottom edge of the table while lying on its side.
    \item \textit{open the bottle cap}: the bottle cap is rotated from a closed position so that it can be lifted off the bottle.
    \item \textit{insert the bread into the toaster}: the bread is placed inside any of the toaster slots.
    \item \textit{pick up the bowl}: the bowl is lifted at least 10cm off the table.
    \item \textit{move the pan to the left}: the pan is moved to the left by at least 10cm.
    \item \textit{wipe the table with the sponge, while avoiding the plate on the table}: the sponge is moved in any wiping motion (zigzag, circular, etc.) while remaining in contact with the table throughout, and it should not touch the plate.
    \item \textit{draw a circle 10cm wide with its centre at [0.0,0.3,0.0] with the gripper closed}: the gripper is closed and moved in a circular motion 10cm wide with its centre on the table and 30cm directly in front of the base of the robot.
    \item \textit{unplug the charger}: the charging block is completely removed from the extension plug socket.
    \item \textit{take out tissue from the dispenser}: the tissue is completely removed from the dispenser.
    \item \textit{lower the brightness of the lamp}: the dimmer switch is rotated anticlockwise.
    \item \textit{hang the towel on the rack}: the towel is placed stably on the rack without touching the table.
\end{itemize}

\end{document}